\documentclass[fleqn]{llncs}

\usepackage[cmex10]{amsmath}
\usepackage{amssymb}
\usepackage{mathtools}
\usepackage{bm}
\usepackage{hyperref}

\newcommand{\mat}[1]{\boldsymbol{{#1}}}

\renewcommand{\vec}[1]{\boldsymbol{{#1}}}

\newcommand{\nrm}[1]{\bigl \lVert #1 \bigr \rVert^2}

\newcommand{\dsq}[2]{\bigl \lVert #1 - #2 \bigr \rVert^2}
\newcommand{\DSQ}[2]{\Bigl \lVert #1 - #2 \Bigr \rVert^2}

\newcommand{\tr}[1]{\operatorname{tr} \bigl [ #1 \bigr ]}
\newcommand{\TR}[1]{\operatorname{tr} \Bigl [ #1 \Bigr ]}

\hypersetup{%
  pdftitle={k-Means Clustering Is Matrix Factorization},
  pdfauthor={Christian Bauckhage},
  pdfsubject={data science, cluster analysis, latent factor modeling},
  pdfkeywords={k-means clustering, matrix factorization},
  colorlinks=true,
  urlcolor=blue,
  citecolor=red,
  bookmarks=false}

\title{$k$-Means Clustering Is Matrix Factorization}

\author{Christian Bauckhage}

\institute{%
  B-IT, University of Bonn, Bonn, Germany \\
  Fraunhofer IAIS, Sankt Augustin, Germany \\
  \email{http://mmprec.iais.fraunhofer.de/bauckhage.html}}

\begin{document}

\maketitle

\begin{abstract}
We show that the objective function of conventional $k$-means clustering can be expressed as the Frobenius norm of the difference of a data matrix and a low rank approximation of that data matrix. In short, we show that $k$-means clustering is a matrix factorization problem. These notes are meant as a reference and intended to provide a guided tour towards a result that is often mentioned but seldom made explicit in the literature.
\end{abstract}

\section{Introduction}

The $k$-means procedure is one of the most popular techniques to cluster a data set $X \subset \mathbb{R}^m$ into subsets $C_1, \ldots, C_k$. The underlying ideas are intuitive and simple and most theoretical properties of $k$-means clustering are well established text book material \cite{MacKay2003-ITI,Hastie2001-TEO}.

In this note, we are concerned with an aspect of $k$-means clustering that is arguably less well known and somewhat under-appreciated. Over the past years, several authors have pointed out that $k$-means clustering can be understood as a constrained matrix factorization problem \cite{Ding2005-OTE,Gaussier2005-RBP,Kim2008-SNN,Arora2013-SBC,Bauckhage2015-CGB}. However, reading these or related texts, it appears as if most authors consider this fact self explanatory and hardly discuss it in detail. Since this may confuse less experienced readers, our goal in this note is to rigorously establish the following equalities for the objective function of hard $k$-means clustering
\begin{equation}
  \label{eq:objective}
  \sum_{i=1}^k \sum_{j=1}^n \, z_{ij} \, \dsq{\vec{x}_j}{\vec{\mu}_i}
  = \DSQ{\mat{X}}{\mat{M} \mat{Z}}
  = \DSQ{\mat{X}}{\mat{X} \mat{Z}^T \bigl(\mat{Z} \mat{Z}^T\bigr)^{-1}\mat{Z}}
\end{equation}
where
\begin{align}
\mat{X} & \in \mathbb{R}^{m \times n} \text{ is a matrix of data vectors } \vec{x}_j \in \mathbb{R}^m  \\
\mat{M} & \in \mathbb{R}^{m \times k} \text{ is a matrix of cluster centroids }  \vec{\mu}_i \in \mathbb{R}^m  \\
\mat{Z} & \in \mathbb{R}^{k \times n} \; \text{ is a matrix of binary indicator variables such that} \notag \\
& z_{ij} =
\begin{cases}
1, \text{ if } \vec{x}_j \in C_i  \\
0, \text{ otherwise.}
\end{cases}
\label{eq:matZ}
\end{align}

\section{Notation and Preliminaries}

Throughout, we write $\vec{x}_j$ to denote $j$-th column vector of a matrix $\mat{X}$. To refer to the $(l,j)$ element of a matrix $\mat{X}$, we either write $x_{lj}$ or $\bigl( \mat{X} \bigr)_{lj}$.

The Euclidean norm of a vector will be written as $\lVert \vec{x} \rVert$ and the Frobenius norm of a matrix as $\lVert \mat{X} \rVert$.

Regarding the squared Frobenius norm of a matrix, we recall the following properties
\begin{equation}
\label{eq:norm}
\nrm{\mat{X}} = \sum_{l, j} \, x_{lj}^2 = \sum_j \, \nrm{\vec{x}_j} = \sum_j \, \vec{x}_j^T \vec{x}_j
= \sum_j \, \bigl( \mat{X}^T \mat{X} \bigr)_{jj} = \tr{\mat{X}^T \mat{X}}
\end{equation}

Finally, subscripts or summation indices $i$ will be understood to range from $1$ to $k$ (the number of clusters), subscripts or summation indices $j$ will range from $1$ up to $n$ (the number of data vectors), and subscripts or summation indices $l$ will be used to expand inner products between vectors or rows and columns of matrices.

\section{Step by Step Derivation of \eqref{eq:objective}}

To substantiate the claim in \eqref{eq:objective}, we first point out several peculiar properties of the binary indicator matrix $\mat{Z}$ in \eqref{eq:matZ}.

If the clusters $C_1, \ldots C_k$ have distinct cluster centroids $\vec{\mu}_1, \ldots, \vec{\mu}_k$, each of the $j$ columns of $\mat{Z}$ will contain a single $1$ and $k-1$ elements that are $0$. Accordingly, the columns of $\mat{Z}$ will sum to one
\begin{equation}
\label{eq:colsum}
\sum_i \, z_{ij} = 1
\end{equation}
and its row sums will indicate the number elements per cluster
\begin{equation}
\label{eq:rowsum}
\sum_j \, z_{ij} = n_i = \lvert C_i \rvert.
\end{equation}

Moreover, since $z_{ij} \in \{0,1\}$ and each column of $\mat{Z}$ only contains a single $1$, the rows of $\mat{Z}$ are pairwise perpendicular because
\begin{equation}
z_{ij} \, z_{i'j} =
\begin{cases}
1, & \text{ if } i = i' \\
0, & \text{ otherwise}
\end{cases}
\end{equation}
which is then to say that the matrix $\mat{Z} \mat{Z}^T$ is a diagonal matrix where
\begin{equation}
\bigl( \mat{Z} \mat{Z}^T \bigr)_{ii'}
= \sum_j \, \bigl( \mat{Z} \bigr)_{ij} \bigl( \mat{Z}^T \bigr)_{j i'}
= \sum_j \, z_{ij} \, z_{i'j}
=
\begin{cases}
n_i, & \text{ if } i = i' \\
0, & \text{ otherwise.}
\end{cases}
\end{equation}

Having familiarized ourselves with these properties of the indicator matrix, we are now positioned to establish the equalities in \eqref{eq:objective} which we will do in a step by step manner.

\subsection{Step 1: Expanding the expression on the left of \eqref{eq:objective}}

We begin our derivation by expanding the conventional $k$-means objective function on the left of \eqref{eq:objective}. For this expression, we have
\begin{align}
\label{eq:expandl}
\sum_{i, j} \, z_{ij} \, \dsq{\vec{x}_j}{\vec{\mu}_i}
& = \sum_{i, j} \, z_{ij} \, \bigl( \vec{x}_j^T \vec{x}_j - 2 \vec{x}_j^T \vec{\mu}_i + \vec{\mu}_i^T \vec{\mu}_i\bigr) \notag \\
& = \underbrace{\sum_{i, j} \, z_{ij} \, \vec{x}_j^T \vec{x}_j}_{T_1}
- 2 \underbrace{\sum_{i, j} \, z_{ij} \, \vec{x}_j^T \vec{\mu}_i}_{T_2}
+   \underbrace{\sum_{i, j} \, z_{ij} \, \vec{\mu}_i^T \vec{\mu}_i}_{T_3} .
\end{align}

This expansion leads to further insights, if we examine the three terms $T_1$, $T_2$, and $T_3$ one by one. First of all, we find
\begin{align}
T_1 = \sum_{i, j} \, z_{ij} \, \vec{x}_j^T \vec{x}_j
& = \sum_{i, j} \, z_{ij} \, \nrm{\vec{x}_j} \\
& = \sum_j \, \nrm{\vec{x}_j} \\
& = \tr{\mat{X}^T \mat{X}}
\end{align}
where we made use of \eqref{eq:colsum} and \eqref{eq:norm}. Second of all, we observe
\begin{align}
T_2 = \sum_{i, j} \, z_{ij} \, \vec{x}_j^T \vec{\mu}_i
& = \sum_{i, j} \, z_{ij} \, \sum_l \, x_{lj} \, \mu_{li} \\
& = \sum_{j, l} \, x_{lj} \, \sum_{i} \, \mu_{li} \, z_{ij} \\
& = \sum_{j, l} \, x_{lj} \, \bigl( \mat{M} \mat{Z} \bigr)_{lj} \\
& = \sum_j \sum_l \, \bigl( \mat{X}^T \bigr)_{jl} \, \bigl( \mat{M} \mat{Z} \bigr)_{lj} \\
& = \sum_j \, \bigl( \mat{X}^T \mat{M} \mat{Z} \bigr)_{jj} \\
& = \tr{\mat{X}^T \mat{M} \mat{Z}}
\end{align}
Third of all, we note that
\begin{align}
T_3 = \sum_{i, j} \, z_{ij} \, \vec{\mu}_i^T \vec{\mu}_i
& = \sum_{i, j} \, z_{ij} \, \nrm{\vec{\mu}_i} \\
& = \sum_i \, \nrm{\vec{\mu}_i} \, n_i
\end{align}
where we applied \eqref{eq:rowsum}.

\subsection{Step 2: Expanding the expression in the middle of \eqref{eq:objective}}

Next, we look at the second expression in \eqref{eq:objective}. As a squared Frobenius norm of a matrix difference, it can be written as
\begin{align}
\label{eq:expandm}
\DSQ{\mat{X}}{\mat{M} \mat{Z}}
& = \TR{\bigl( \mat{X} - \mat{M} \mat{Z} \bigr)^T \bigl( \mat{X} - \mat{M} \mat{Z} \bigr)} \notag \\
& = \underbrace{\tr{\mat{X}^T \mat{X}}}_{T_4}
- 2 \underbrace{\tr{\mat{X}^T \mat{M} \mat{Z}}}_{T_5}
+   \underbrace{\tr{\mat{Z}^T \mat{M}^T \mat{M} \mat{Z}}}_{T_6}
\end{align}

Given our earlier results, we immediately recognize that $T_1 = T_4$ and $T_2 = T_5$. Thus, to establish that \eqref{eq:expandl} and \eqref{eq:expandm} are indeed equivalent, it remains to verify whether $T_3 = T_6$?

Regarding $T_6$, we note that, because of the cyclic permutation invariance of the trace operator, we have
\begin{equation}
\tr{\mat{Z}^T \mat{M}^T \mat{M} \mat{Z}} = \tr{\mat{M}^T \mat{M} \mat{Z} \mat{Z}^T}.
\end{equation}
We also note that
\begin{align}
\tr{\mat{M}^T \mat{M} \mat{Z} \mat{Z}^T}
& = \sum_i \, \bigl( \mat{M}^T \mat{M} \mat{Z} \mat{Z}^T \bigr)_{ii} \\
& = \sum_i \sum_l \, \bigl( \mat{M}^T \mat{M} \bigr)_{il} \bigl( \mat{Z} \mat{Z}^T \bigr)_{li} \\
& = \sum_i \, \bigl( \mat{M}^T \mat{M} \bigr)_{ii} \bigl( \mat{Z} \mat{Z}^T \bigr)_{ii} \\
& = \sum_i \, \nrm{\vec{\mu}_i} \, n_i
\end{align}
where we used the fact that $\mat{Z} \mat{Z}^T$ is diagonal. This result, however, shows that $T_3 = T_6$ and, consequently, that \eqref{eq:expandl} and \eqref{eq:expandm} really are equivalent.

\subsection{Step 3: Eliminating matrix $\mat{M}$}

Finally, to establish the equality on the right of \eqref{eq:objective} we ask for the matrix $\mat{M}$ that, for a given $\mat{Z}$, would minimize $\dsq{\mat{X}}{\mat{M} \mat{Z}}$. To this end, we consider
\begin{align}
\frac{\partial}{ \partial \mat{M}} \DSQ{\mat{X}}{\mat{M} \mat{Z}}
& = \frac{\partial}{ \partial \mat{M}} \left[ \tr{\mat{X}^T \mat{X}} - 2 \, \tr{\mat{X}^T \mat{M} \mat{Z}} + \tr{\mat{Z}^T \mat{M}^T \mat{M} \mat{Z}} \right] \notag \\
& = 2 \bigl( \mat{M} \mat{Z} \mat{Z}^T - \mat{X} \mat{Z}^T \bigr)
\end{align}
which, upon equation to $\mat{0}$, leads to
\begin{equation}
\mat{M} = \mat{X} \mat{Z}^T \bigl( \mat{Z} \mat{Z}^T \bigr)^{-1}
\end{equation}
which beautifully reflects the fact that each of the $k$-means cluster centroids $\vec{\mu}_i$ coincides with the mean of the corresponding cluster $C_i$, namely
\begin{equation}
\vec{\mu}_i = \frac{\sum_j z_{ij} \, \vec{x}_j}{\sum_j z_{ij}} = \frac{1}{n_i} \sum_{\vec{x}_j \in C_i} \vec{x}_j.
\end{equation}

\section{Conclusion}

Using tedious yet straightforward algebra, we have shown the the problem of hard $k$-means clustering can be understood as the following constrained matrix factorization problem
\begin{equation*}
\begin{aligned}
& \min_{\mat{Z}} && \DSQ{\mat{X}}{\mat{X} \mat{Z}^T \bigl(\mat{Z} \mat{Z}^T\bigr)^{-1}\mat{Z}} \\[1ex]
& \text{ s.t.} && \; z_{ij} \in \{0, 1\} \\
&&& \sum_j \, z_{ij} = 1
\end{aligned}
\end{equation*}

\bibliographystyle{splncs}
\bibliography{literature}

\begin{thebibliography}{1}

\bibitem{MacKay2003-ITI}
MacKay, D.:
\newblock {Information Theory, Inference, \& Learning Algorithms}.
\newblock Cambridge University Press (2003)

\bibitem{Hastie2001-TEO}
Hastie, T., Tibshirani, R., Friedman, J.:
\newblock {The Elements of Statistical Learning}.
\newblock Springer (2001)

\bibitem{Ding2005-OTE}
Ding, C., He, X., Simon, H.:
\newblock {On the Equivalence of Nonnegative Matrix Factorization and Spectral
  Clustering}.
\newblock In: Proc. SDM, SIAM (2005)

\bibitem{Gaussier2005-RBP}
Gaussier, E., Goutte, C.:
\newblock {Relations between PLSA and NMF and Implications}.
\newblock In: Proc. SIGIR, ACM (2005)

\bibitem{Kim2008-SNN}
Kim, J., Park, H.:
\newblock {Sparse Nonnegative Matrix Factorization for Clustering}.
\newblock Technical Report GT-CSE-08-01, Georgia Institute of Technology (2008)

\bibitem{Arora2013-SBC}
Arora, R., Gupta, M., Kapila, A., Fazel, M.:
\newblock {Similarity-based Clustering by Left-Stochastic Matrix
  Factorization}.
\newblock J. of Machine Learning Research \textbf{14}(Jul.) (2013)

\bibitem{Bauckhage2015-CGB}
Bauckhage, C., Drachen, A., Sifa, R.:
\newblock {Clustering Game Behavior Data}.
\newblock IEEE Trans. on Computational Intelligence and AI in Games
  \textbf{7}(3) (2015)

\end{thebibliography}

\end{document}